\documentclass[a4paper, 10pt, conference]{ieeeconf}      

\IEEEoverridecommandlockouts                              
\usepackage[utf8]{inputenc}
\usepackage[T1]{fontenc}
\usepackage[pdftex]{graphicx}
\usepackage{graphicx}
\usepackage{tikz}
\usepackage{subcaption}
\graphicspath{{../Figures/}}
\usepackage{multirow}
\usepackage{float}
\usepackage{graphics} 
\usepackage{multicol}
\usepackage{subcaption}
\usepackage{color, colortbl}
\usepackage[]{algorithm2e}
\usepackage{amsmath} 
\usepackage{array}  
   
\usepackage{bm} 

\title{\LARGE \bf
Semantic sensor fusion: from camera to sparse lidar information
}

\author{Julie Stephany Berrio $^{1}$, Mao Shan $^{1}$, Stewart Worrall $^{1}$, James Ward $^{1}$, Eduardo Nebot$^{1}$
\thanks{$^{1}$J. Berrio, M. Shan, S. Worrall, J. Ward, , E. Nebot  are with the Australian Centre for Field Robotics (ACFR) at the University of Sydney (NSW, Australia).
       E-mails: {\tt\small \{j.berrio, m.shan, s.worrall, j.ward, e.nebot\}@acfr.usyd.edu.au}}
}

\begin{document}

\maketitle
\thispagestyle{empty}
\pagestyle{empty}

\begin{abstract}

To navigate through urban roads, an automated vehicle must be able to perceive and recognize objects in a three-dimensional environment. A high-level contextual understanding of the surroundings is necessary to plan and execute accurate driving maneuvers.
This paper presents an approach to fuse different sensory information, Light Detection and Ranging (lidar) scans and camera images.
The output of a convolutional neural network (CNN) is used as classifier to obtain the labels of the environment.
The transference of semantic information between the labelled image and the lidar point cloud is performed in four steps: initially, we use heuristic methods to associate probabilities to all the semantic classes contained in the labelled images.  
Then, the lidar points are corrected to compensate for the vehicle's motion given the difference between the timestamps of each lidar scan and camera image. In a third step, we calculate the pixel coordinate for the corresponding camera image. In the last step we perform the transfer of semantic information from the heuristic probability images to the lidar frame, while removing the lidar information that is not visible to the camera.
We tested our approach in the Usyd Dataset \cite{usyd_dataset}, obtaining qualitative and quantitative results that demonstrate the validity of our probabilistic sensory fusion approach.

\end{abstract}

\section{INTRODUCTION}

Getting a sufficiently descriptive representation of the current state of the world in an urban environment has always been a challenge. Today, with the recent boom in self-driving car applications, more information about the shape, location and type of nearby objects is required to assure not only accuracy/reliability in localization but also safe and optimal path planning and navigation. Semantic meaning of the objects around the vehicle is one of the most significant pieces of information necessary in the decision making process. 

\begin{figure}[!h]
\vspace{3mm}
\centerline{
\includegraphics[width=0.99\columnwidth]{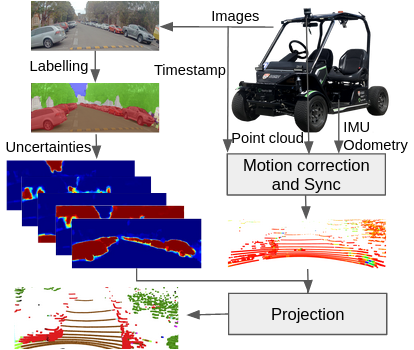}
}
\caption{Pipeline of the process. }
\label{fig:uncertainties}
\end{figure}

The classification techniques to label objects in the surrounding area relies on both the sensors located on the vehicle, and the computational power available for this task. Cameras and lidars are often used to accomplish the task of understanding the environment. Cameras are affordable, have low power consumption and contain texture and colour data, which makes them a suitable sensor to be used for object classification \cite{ramos_2015}. 
In this paper we use the captured images as the input for a trained CNN \cite{wie_2019} which outputs images with a semantic representation of the objects in the image. We have developed a heuristic method to associate class uncertainty to each pixel by modifying the CNN's final softmax function based on the label distribution within the original image's superpixels. 

From only monocular images it is not possible to measure directly the 3D location of detected objects. To overcome this restriction, lidar sensors have been widely used in the autonomous vehicles research. Typically, low-cost lidars do not have sufficient resolution, and are unable to provide color information. 

Through camera-lidar sensor fusion, we are able to transfer relevant data from the camera to lidar and vice versa, providing a better understanding of the structure of the surroundings \cite{Hsiang_2016}. A proper lidar-camera data fusion approach must address the following three problems: Accurate calibration for intrinsic camera and extrinsic lidar-camera parameters, sensor synchronization and motion correction, and handling occlusion caused by the different point of view of each sensor.

Different papers have addressed sensor fusion between cameras and lidars, using only the geometric relationship between a pin-hole camera and the lidar. In \cite{2015_li}, authors presented an indoor scene construction technique using a 2D lidar and a digital camera. Both sensors were mounted rigidly, and the sensor fusion was performed using the extrinsic calibration parameters. Later, Bybee et. al, in \cite{Bybee_2019} presented a bundle adjustment technique to fuse low-cost lidar information and camera data to create terrain models. A multi-object tracking technique which rigidly fuses object proposals across sensors is explained in \cite{Rangesh_2019}.

Schneider et. al. in \cite{Schneider_2010} addressed the synchronization and motion correction problem by triggering the cameras to capture the image when the laser-beams are in the images field of view. The scan points were translated based on the the vehicle movement. In \cite{8317846_supersensor}, the synchronization timestamp is chosen to coincide with the timestamp of the most recent camera frame, then transforming each point using the ego-motion transform matrix of the vehicle.

The most common solution to the occlusion problem corresponds to segment the point cloud. A 2D convex hull is computed for every cluster in the image frame, then an occlusion check is performed to takes into account the depth of each cluster to select which part of the point cloud is not occluded to the camera view \cite{Schneider_2010}. In \cite{Pintus_2011}, authors presented an approach where a cone is placed along the projection line for every 3D point, with the condition that it does not intersect any other point. The aperture of the cone has to be larger than a threshold to set the point as visible.
These techniques work well for a dense point cloud which allows the segmentation algorithm to provide consistent results. Nevertheless, for sparse point clouds such as the type provided by the 16 beam lidar used in this work, only objects that are very close to the vehicle can be reliably segmented. This is not very useful in an urban type environments.

\begin{figure}[!t]
\vspace{3mm}
\centerline{
\includegraphics[width=0.9\columnwidth]{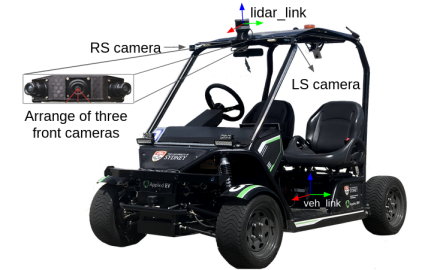}
}
\caption{Experimental platform equipped with five cameras (two side cameras right-side RS and left-side LS, and an arrangement of three front facing cameras) running at 30 Hz, one velodyne VLP-16 lidar (16 laser beams) with a frequency of 10 Hz, one IMU and wheel encoders at a rate of 100 Hz. }
\label{fig:platform}
\end{figure}

In contrast of previous approaches, this paper proposes a novel pipeline for the fusion of semantically labelled images and lidar point clouds which deals with both motion distortion and the occlusion problem. We tested our approach in the Usyd Dataset \cite{usyd_dataset}, \cite{USYD_Segmentation_2019}, which was obtained with the vehicle platform shown in Fig.\ref{fig:platform}. 

In the next section, we explain the details of the methodology to associate uncertainties to the labelled image and project the resulting information into the point cloud. The algorithms also include the correction for vehicle motion and a process for handling occlusions. Experiments and outcomes are presented in section III.

\section{METHODOLOGY}

An E-net CNN which has been fine tuned with locally annotated images and data augmentation techniques \cite{wie_2019} is adopted as semantic classifier. This model is used to generate the labels, and a heuristic value for measurement uncertainty. 

The complete lidar scan, divided in packets, is translated based on the motion of the vehicle and the cameras's timestamp. Following this, we calculate the corresponding image coordinate for each measurement point. 

\begin{figure*}[h!]
\vspace{3mm}
\centering

\begin{subfigure}[]{0.48\columnwidth}
\centering
	\includegraphics[width=\columnwidth, height=2.4cm]{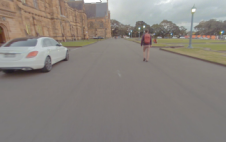}
    \caption{Original image}
    \label{sub_a}
    \end{subfigure}
\begin{subfigure}[]{0.48\columnwidth}
\centering
	\includegraphics[width=\columnwidth, height=2.4cm]{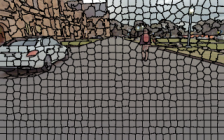}
    \caption{Superpixel clustering}
    \label{sub_b}
    \end{subfigure}
\begin{subfigure}[]{0.48\columnwidth}
\centering
	\includegraphics[width=\columnwidth, height=2.4cm]{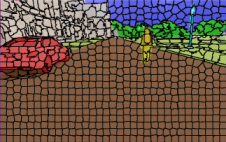}
    \caption{Semantic labels and Sp}
    \label{sub_c}
    \end{subfigure}
\begin{subfigure}[]{0.48\columnwidth}
\centering
	\includegraphics[width=\columnwidth, height=2.5cm]{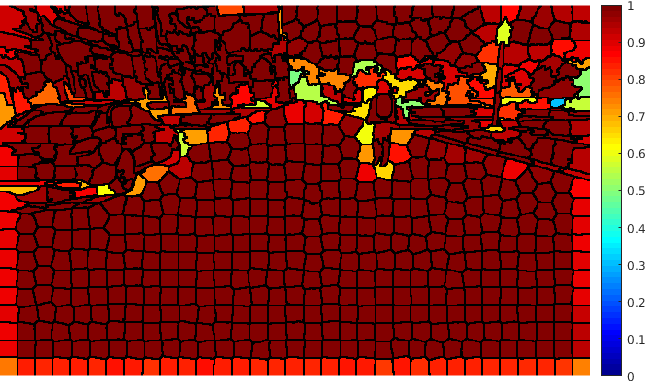}
    \caption{$\%$ of the label mode per Sp}
    \label{sub_d}
    \end{subfigure}

\caption{\small Uncertainty association process. The raw image captured by the camera is shown in \ref{sub_a}. \ref{sub_b} shows the clustering performed by the SLIC algorithm to the original image. In \ref{sub_c} the CNN's semantic segmentation result is overlaid with the superpixel segmentation. Here the colour code is: red for vehicles, white for buildings, brown for roads, green for vegetation, blue for sky, lime is for undrivable roads, yellow for pedestrians and riders, cyan for poles, gray for fence and purple for unlabeled pixels, for a total of 12 classes. \ref{sub_d} displays the result of $spp_k$ within the superpixels, the color bar represents percentage of the most common label within the superpixel.}
\label{fig:proceso}
\end{figure*}

\begin{figure*}[t!]
\vspace{3mm}
\centering

\begin{subfigure}[]{0.32\columnwidth}
\centering
	\includegraphics[trim={0 0 2cm 0},clip,width=\columnwidth]{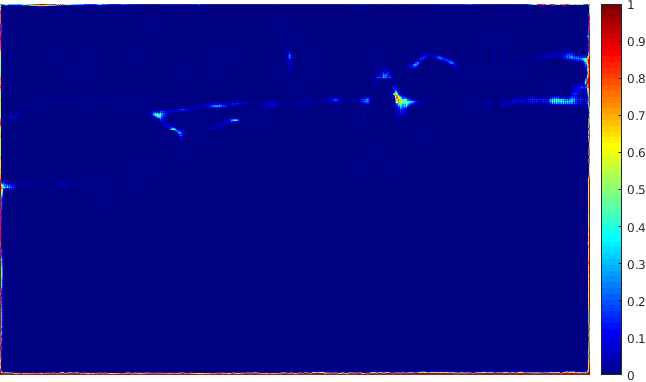}
    \caption{Unlabled}
    \label{sub_a1}
    \end{subfigure}
\begin{subfigure}[]{0.32\columnwidth}
\centering
	\includegraphics[trim={0 0 2cm 0},clip,width=\columnwidth]{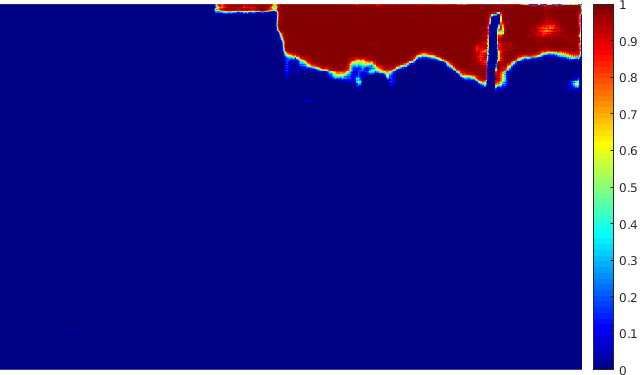}
    \caption{Sky}
    \label{sub_b1}
    \end{subfigure}
\begin{subfigure}[]{0.32\columnwidth}
\centering
	\includegraphics[trim={0 0 2cm 0},clip,width=\columnwidth]{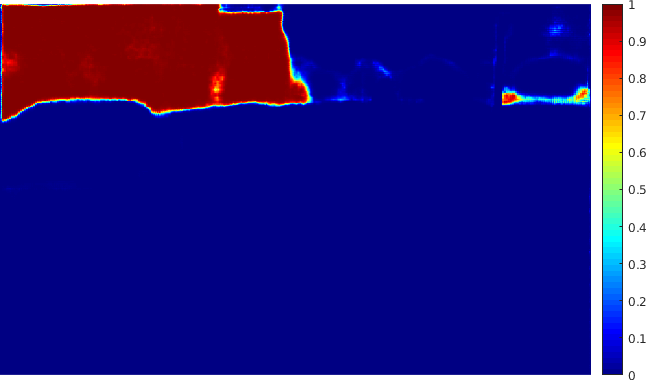}
    \caption{Building}
    \label{sub_c1}
    \end{subfigure}
\begin{subfigure}[]{0.32\columnwidth}
\centering
	\includegraphics[trim={0 0 2cm 0},clip,width=\columnwidth]{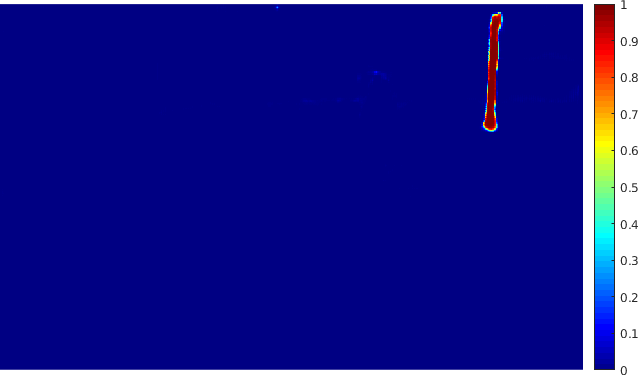}
    \caption{Pole}
    \label{sub_d1}
    \end{subfigure}
\begin{subfigure}[]{0.32\columnwidth}
\centering
	\includegraphics[trim={0 0 2cm 0},clip,width=\columnwidth]{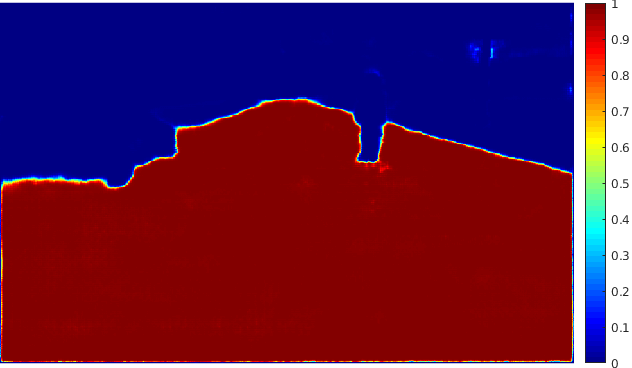}
    \caption{Road}
    \label{sub_e1}
    \end{subfigure}
\begin{subfigure}[]{0.32\columnwidth}
\centering
	\includegraphics[trim={0 0 2cm 0},clip,width=\columnwidth]{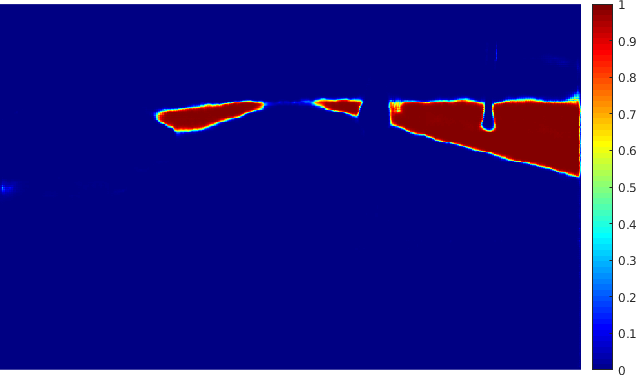}
    \caption{U. Road}
    \label{sub_f1}
    \end{subfigure}

\begin{subfigure}[]{0.32\columnwidth}
\centering
	\includegraphics[trim={0 0 2cm 0},clip,width=\columnwidth]{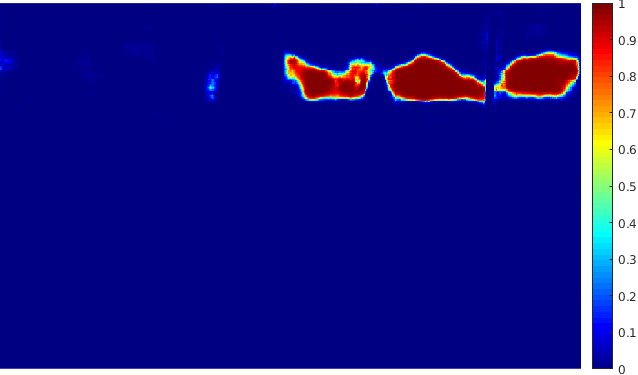}
    \caption{Vegetation}
    \label{sub_g1}
    \end{subfigure}
\begin{subfigure}[]{0.32\columnwidth}
\centering
	\includegraphics[trim={0 0 2cm 0},clip,width=\columnwidth]{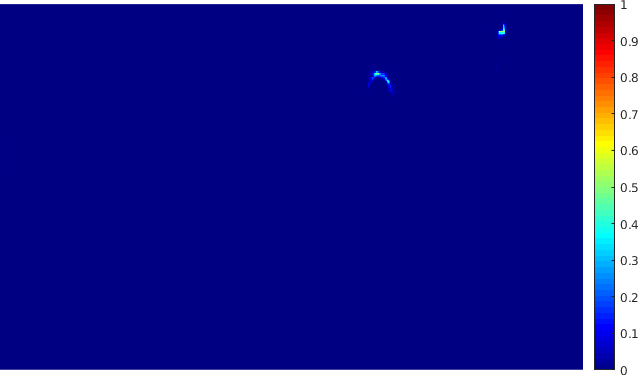}
    \caption{Sign}
    \label{sub_h1}
    \end{subfigure}
\begin{subfigure}[]{0.32\columnwidth}
\centering
	\includegraphics[trim={0 0 2cm 0},clip,width=\columnwidth]{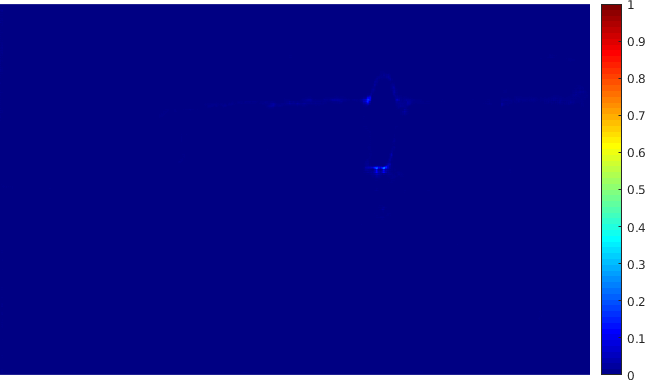}
    \caption{Fence}
    \label{sub_i1}
    \end{subfigure}
\begin{subfigure}[]{0.32\columnwidth}
\centering
	\includegraphics[trim={0 0 2cm 0},clip,width=\columnwidth]{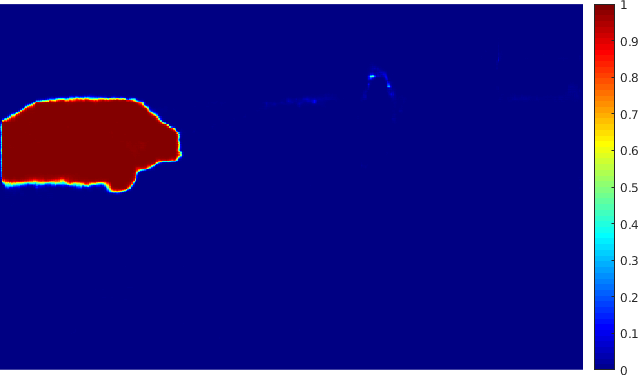}
    \caption{Vehicle}
    \label{sub_j1}
    \end{subfigure}
\begin{subfigure}[]{0.32\columnwidth}
\centering
	\includegraphics[trim={0 0 2cm 0},clip,width=\columnwidth]{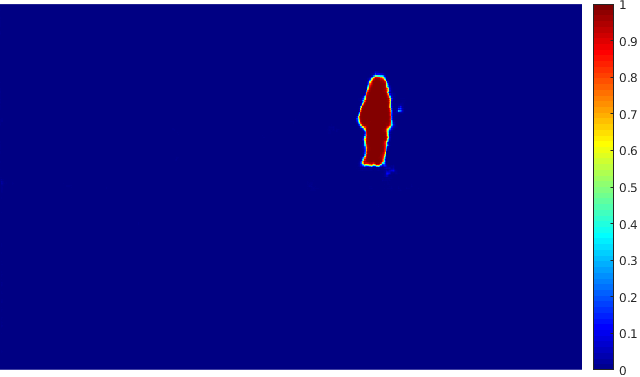}
    \caption{Pedestrian}
    \label{sub_k1}
    \end{subfigure}
\begin{subfigure}[]{0.32\columnwidth}
\centering
	\includegraphics[trim={0 0 2cm 0},clip,width=\columnwidth]{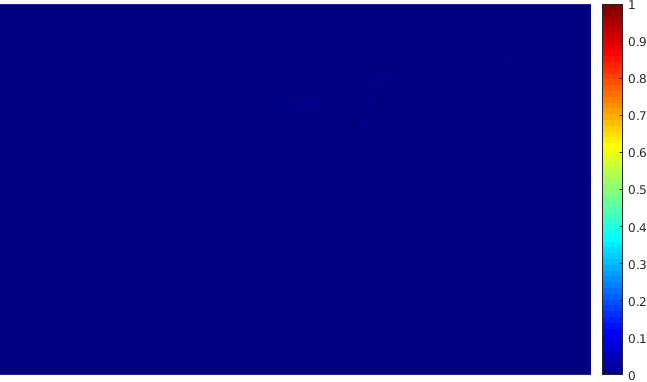}
    \caption{Rider}
    \label{sub_l1}
    \end{subfigure}

\begin{subfigure}[]{0.32\columnwidth}
\centering
	\includegraphics[trim={0 0 2cm 0},clip,width=\columnwidth]{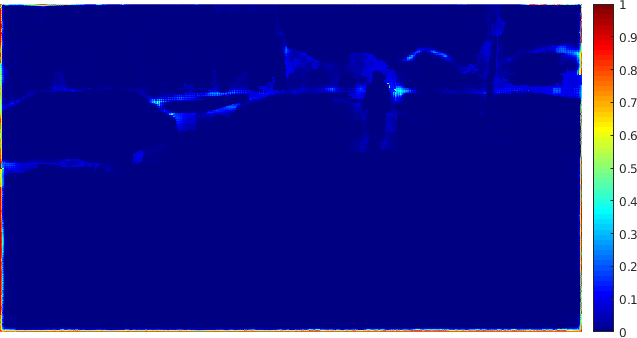}
    \caption{Unlabeled}
    \label{sub_m1}
    \end{subfigure}
\begin{subfigure}[]{0.32\columnwidth}
\centering
	\includegraphics[trim={0 0 2cm 0},clip,width=\columnwidth]{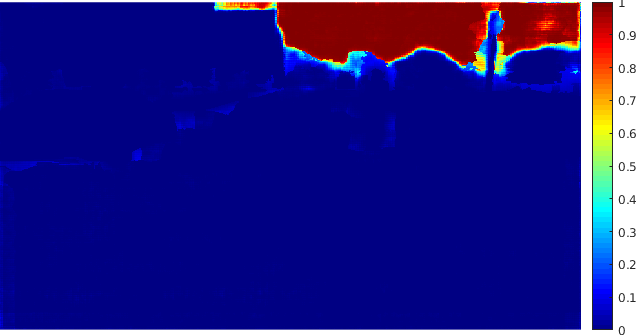}
    \caption{Sky}
    \label{sub_n1}
    \end{subfigure}
\begin{subfigure}[]{0.32\columnwidth}
\centering
	\includegraphics[trim={0 0 2cm 0},clip,width=\columnwidth]{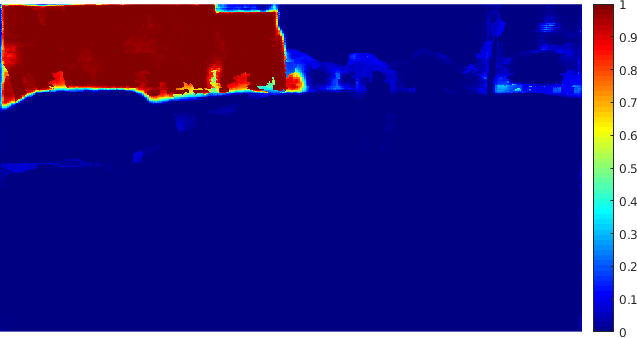}
    \caption{Building}
    \label{sub_o1}
    \end{subfigure}
\begin{subfigure}[]{0.32\columnwidth}
\centering
	\includegraphics[trim={0 0 2cm 0},clip,width=\columnwidth]{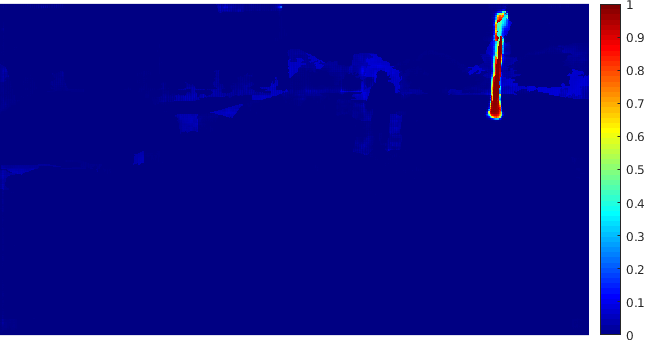}
    \caption{Pole}
    \label{sub_p1}
    \end{subfigure}
\begin{subfigure}[]{0.32\columnwidth}
\centering
	\includegraphics[trim={0 0 2cm 0},clip,width=\columnwidth]{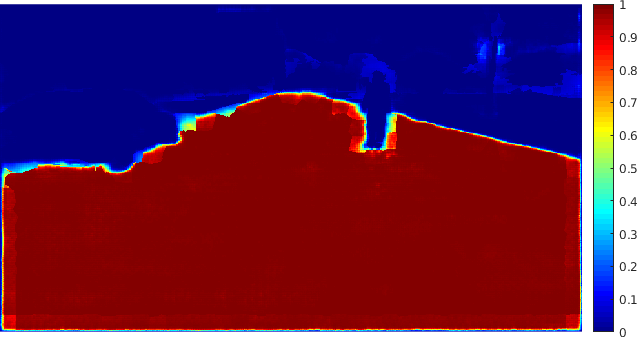}
    \caption{Road}
    \label{sub_q1}
    \end{subfigure}
\begin{subfigure}[]{0.32\columnwidth}
\centering
	\includegraphics[trim={0 0 2cm 0},clip,width=\columnwidth]{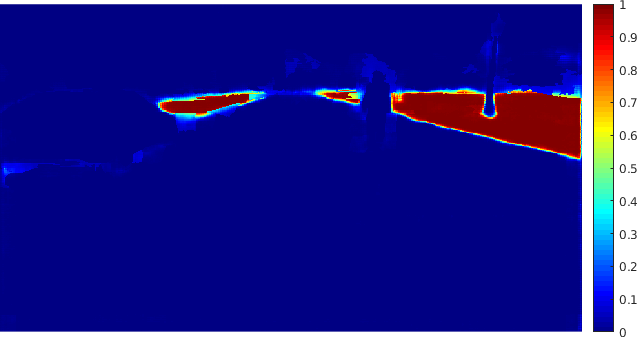}
    \caption{U. Road}
    \label{sub_r1}
    \end{subfigure}

\begin{subfigure}[]{0.32\columnwidth}
\centering
	\includegraphics[trim={0 0 2cm 0},clip,width=\columnwidth]{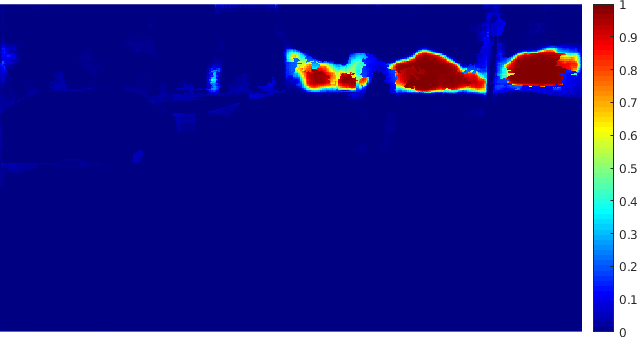}
    \caption{Vegetation}
    \label{sub_s1}
    \end{subfigure}
\begin{subfigure}[]{0.32\columnwidth}
\centering
	\includegraphics[trim={0 0 2cm 0},clip,width=\columnwidth]{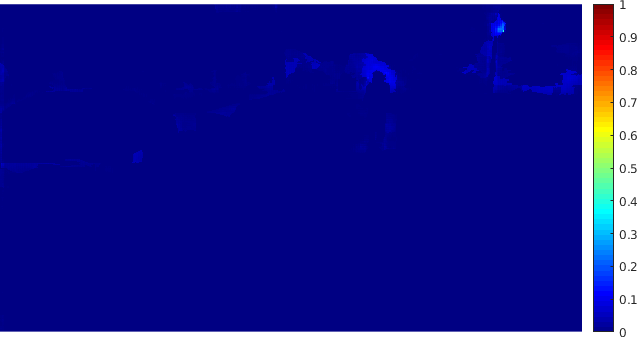}
    \caption{Sign}
    \label{sub_t1}
    \end{subfigure}
\begin{subfigure}[]{0.32\columnwidth}
\centering
	\includegraphics[trim={0 0 2cm 0},clip,width=\columnwidth]{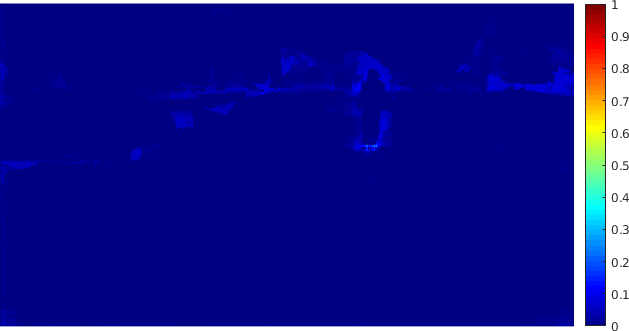}
    \caption{Fence}
    \label{sub_u1}
    \end{subfigure}
\begin{subfigure}[]{0.32\columnwidth}
\centering
	\includegraphics[trim={0 0 2cm 0},clip,width=\columnwidth]{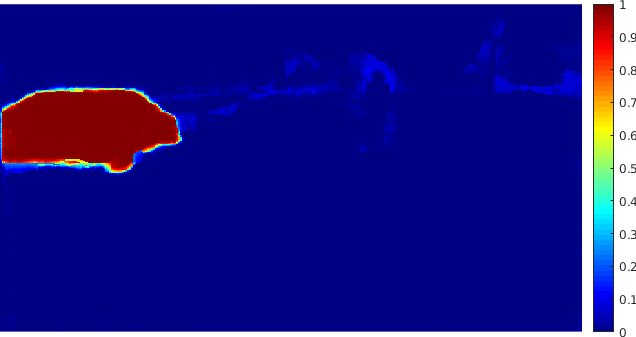}
    \caption{Vehicle}
    \label{sub_v1}
    \end{subfigure}
\begin{subfigure}[]{0.32\columnwidth}
\centering
	\includegraphics[trim={0 0 2cm 0},clip,width=\columnwidth]{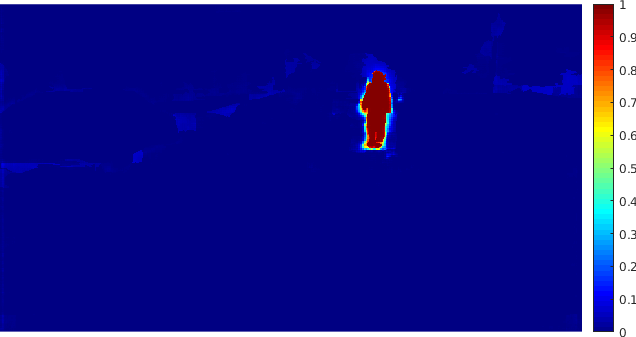}
    \caption{Pedestrian}
    \label{sub_w1}
    \end{subfigure}
\begin{subfigure}[]{0.32\columnwidth}
\centering
	\includegraphics[trim={0 0 2cm 0},clip, width=\columnwidth]{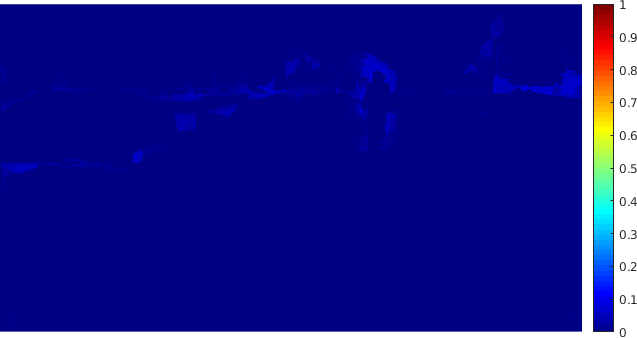}
    \caption{Rider}
    \label{sub_y1}
    \end{subfigure}

\caption{\small Uncertainty association process. Comparison between the original CNN's score maps (first two rows) and the proposed method (last two rows) changing the temperature of the softmax function based on $spp_k$. Dark red represents the highest class probability and dark blue is the lowest class probability.}
\label{fig:proceso1}
\end{figure*}

The projection of the image information to the point cloud is done by applying a masking technique that deals with the occlusion problem, avoiding the projection to reject the lidar points which are not visible from the camera. Each valid 3D point then gets assigned the probabilistic distribution of the semantic classes calculated in an earlier step. The process is explained in this section.

\subsection{Heuristic label probabilities}

The final module of E-NET consists of a bare full convolution which outputs a three dimensional $c$ x $n$ x $m$ feature/activation map, where $n$ and $m$ correspond to the size of the input image, and $c$ is the number of object classes \cite{Paszke2016ENetAD}. 
Each $c$ feature map consists of $c$ activations per pixel $[u,v]$, which represents the unnormalised class score $S_c$ for each label \cite{simonyan14deep}. 

The output unit activation function for the CNN model, which conforms the canonical link, is determined by the softmax function. \cite{10.1007/978-3-642-76153-9_28_softmax}: 

\begin{equation} \label{eq1}
P_c=\frac{\exp({S(c))}}{\sum_{b=1}^{n_c}\exp({S(b)})}
\end{equation}

The softmax function mixes the class scores $S_c$ while satisfying the constraints 
to adopt the interpretation of $P_c$ as a class probability \cite{Bishop:2006:PRM:1162264}:

\begin{equation} \label{eq2}
\sum_{c}P_c = 1, \qquad 0\leq P_c\leq 1
\end{equation}

The final labeled image $L$ is a $n$ x $m$ matrix composed of the class identifier for the label with the highest probability per pixel.

In this paper, we propose a variant of the method for obtaining the labels' probabilities while retaining the CNN's output classification.  This method is based on both the score maps, and the distribution of the labels within the segmented areas of the input image. 
Initially, the input image is divided into different regions by using simple linear iterative clustering (SLIC) \cite{6205760_slic} super-pixel segmentation method \cite{1238308_superpixel} where pixels are grouped according to perceptual characteristics consistency. Given the uniformity of the pixels in a super-pixel, we assume they all belong to a single semantic class.

Then, we calculate the percentage of the predominant label $spp_k$ within the super-pixel $k$ by dividing the number of pixels belonging to this label by the total amount of pixels inside the super-pixel. In an ideal case where the labels and the super-pixels are correctly segmented, prevalent label percentage $spp_k$  would be $1$. Fig. \ref{fig:proceso} shows the result of this process, it is noticeable that the value of $spp_k$ decreases in the superpixels near the object borders.

In some cases, two or more labels are present within a super-pixel, indicating that at least some pixels are incorrectly labelled. This is often observed with the classification of pixels around the edges of objects in the image, which is due to the re-sizing process performed by the CNN model. In this case, we would expect to have a class probability distributed more evenly among two or more labels. Nevertheless, the softmax function described in (\ref{eq1}) generates a strong discrepancy in the selection probability for dissimilar estimated class scores.

We have unified the concepts of predominant label percentage within a super-pixel and the softmax activation function used by the CNN. This is done through a variant definition of the softmax function: 
\begin{equation} \label{eq3}
P_c=\frac{\exp({S(c)/\tau) }}{\sum_{b=1}^{n}\exp({S(b)/\tau })}
\end{equation}

where $\tau$ represents a positive parameter denominated temperature, high temperatures lead to the selection probability to be approximately equiprobable. Instead, low temperatures bring about a greater difference in selection probability for actions that differ in their value estimates \cite{sutton1992reinforcement}.

Our approach is to modulate the temperature of the softmax function per super-pixel based on its label distribution. This is done to obtain more coherent estimated probabilities of the classification process while satisfying the restriction (\ref{eq2}) and maintaining the classification output. The softmax temperature adjustment for the super-pixel $k$ is done by: 

\begin{equation} \label{eq4}
\tau_k=\frac{1}{spp_k^2}
\end{equation}

The softmax temperature per super-pixel is inversely proportional to the square of $spp_k$, so when $spp_k$ is less than $1$, the temperature is raised. This has the effect of flattening the activation function and consequently generating more distributed probabilities. Where $spp_k$ is equal to 1, the class probabilities are identical to those provided by the CNN model as shown in Fig. \ref{fig:proceso1}.

\subsection{Motion correction}  

By moving the vehicle coordinate system for each individual set of lidar measurements (called a packet) and transforming them into the final scan coordinate frame, it is possible to compensate for the changing reference frame of the lidar \cite{Himmelsbach08lidar-based3d}. 
The motion compensation method implemented in this paper synchronizes the lidar packet timestamps with closest camera image timestamp $t_{ref}$ following a similar principle to the method presented in \cite{8317846_supersensor}.

\begin{figure}[!h]
\vspace{3mm}
\centerline{
\includegraphics[width=0.95\columnwidth]{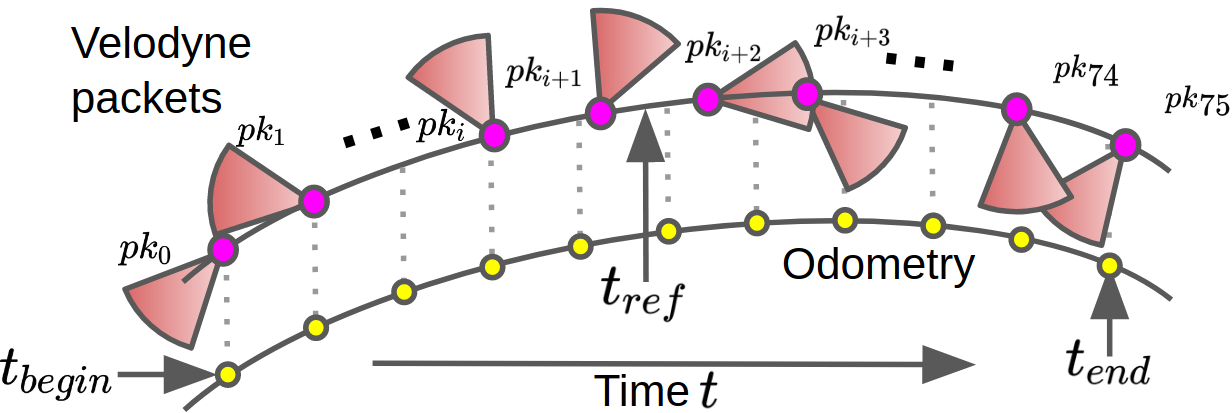}
}
\caption{Lidar point cloud motion correction process.}
\label{fig:time_line}
\end{figure}

Every 3D point packet $i^{th}$ denoted by $pk_{i}$ registered at time $t_i$ is adjusted by eq. (\ref{eq5}). The point is first translated to the vehicle footprint frame by the rigid transform $T^l_{veh}$. Then, a transform is applied to compensate for the displacement resulting from the ego-motion of the vehicle. This transform matrix represents the motion of the vehicle $T_{Ego}$ between the time reference $t_{ref}$ and the odometry reading closest to $t_i$. As a result, each point is converted into the lidar coordinate system.

\begin{equation} \label{eq5}
\widetilde{pk}_{i}=(T^l_{veh})^{-1} \cdot T_{Ego}^{-\Delta _i}\cdot T^l_{veh} \cdot pk_{i}
\end{equation}

Where $\Delta_i$ is defined as the difference between the camera image timestamp or $t_{ref}$ and lidar packet $t_i$. 
In order to get an accurate camera-lidar projection, the motion compensation process is performed for all point cloud packets generated by the lidar. 

\subsection{Image-to-Lidar Projection}

This step takes as an input the adjusted point cloud result of the motion correction performed previously. 
Given the extrinsic calibration between both camera and lidar sensors represented as the transformation matrix $T_{l}^{cn}$, we can translate each 3D point $[\widetilde{x}_{l_i},\widetilde{y}_{l_i},\widetilde{x}_{l_i}]^T$ in the lidar frame to the camera frame $[x_{c_n},y_{c_n},z_{c_n}]^T$ by computing: 

\begin{equation} \label{eq_pc0}
   \begin{bmatrix}
x_{c_n}\\ 
y_{c_n}\\ 
z_{c_n}\\
\end{bmatrix} = T_{l}^{cn} \begin{bmatrix}
\widetilde{x}_{l_i}\\ 
\widetilde{y}_{l_i}\\ 
\widetilde{z}_{l_i}\\
\end{bmatrix}
\end{equation}

Once the point cloud is referenced to the camera frame, we find the corresponding pixel in the image frame by using the camera model and its intrinsic parameters. Initially we make use of the generic pinhole camera-image projection equations which states:

\begin{align}
a &=  \frac{x_{c_n}}{z_{c_n}} & b &=  \frac{y_{c_n}}{z_{c_n}} \label{eq_pc1}
\end{align}

\begin{align}
r &= \sqrt{a^{2}+b^{2}} & \theta &= \textup{atan}(r) \label{eq_pc2}
\end{align}

Since our cameras have fisheye lenses, we need correct for the distortion established by the camera model to find the corresponding pixel in the image \cite{opencv}. The distortion of the lens is calculated as follows: 

\begin{equation} \label{eq_pc3}
   \theta_d = \theta(1+k_1\theta^2+k_2\theta^4+k_3\theta^6+k_4\theta^8)
\end{equation}

where $k_1$, $k_2$, $k_3$ and $k_4$ are the lens' distortion coefficients. We then compute the distorted point coordinates as:

\begin{align}
x' &=  (\theta_d/r)a & y' &=  (\theta_d/r)b
\end{align}

The definite pixel coordinates vector $[u,v]$ in the image frame of the 3D point $[\widetilde{x}_{l_i},\widetilde{y}_{l_i},\widetilde{x}_{l_i}]^T$ can be estimated as:

\begin{align}
u &= f_x*(x'+\alpha y')+c_x & v &= f_y*(y')+c_y
\label{eq_pc4}
\end{align}

where $\alpha$ is the camera's skew coefficient, $[c_x,c_y]$ the principal point offset and $[f_x, f_y]$ are the focal lengths expressed in pixel units. %

\subsubsection{Occlusion handling}

The direct fusion of the camera-lidar information would be appropriate if the coordinate systems of both the camera and the lidar were co-located. Most vehicles are equipped with multiple cameras at different positions to achieve wide coverage.  For this case, the cameras and lidar will see the environment from different vantage points. This can result in a scenario where the lidar can see an object located behind another object that blocks the cameras visibility. In this work we address this problem using a masking technique.

The point cloud in the camera frame $[x_{c_n},  y_{c_n},  z_{c_n}]^T$ is first sorted using a k-dimensional tree to organize the 3D points in ascending order based on their distance to the camera frame origin. The next step is to project each point into the image frame. To cope with the occlusion problem we propose a masking approach, where every projected point generates a mask. The mask is used to reject other more distant points that are within the masked zone of the image. 

\begin{figure}[t!]
\vspace{3mm}
\centering

\begin{subfigure}[]{0.48\columnwidth}
\centering
	\includegraphics[width=0.95\columnwidth]{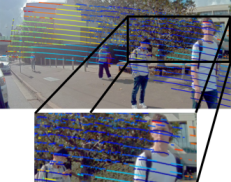}
    \caption{LS camera image.}
    \label{sub1_a}
    \end{subfigure}
\begin{subfigure}[]{0.48\columnwidth}
\centering
	\includegraphics[width=0.95\columnwidth]{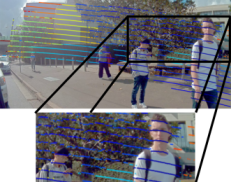}
    \caption{LS masked camera image. }
    \label{sub1_b}
    \end{subfigure}
    
\begin{subfigure}[]{0.48\columnwidth}
\centering
	\includegraphics[width=0.95\columnwidth]{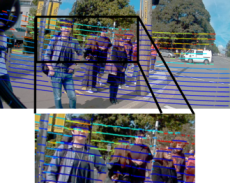}
    \caption{L camera image.}
    \label{sub1_c}
    \end{subfigure}
\begin{subfigure}[]{0.48\columnwidth}
\centering
	\includegraphics[width=0.95\columnwidth]{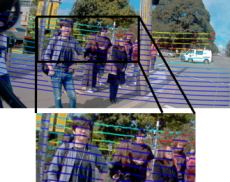}
    \caption{L masked camera image. }
    \label{sub1_d}
    \end{subfigure}
    
\begin{subfigure}[]{0.48\columnwidth}
\centering
	\includegraphics[width=0.9\columnwidth]{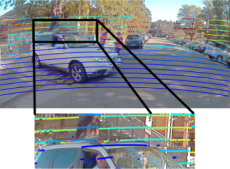}
    \caption{C camera image.}
    \label{sub1_e}
    \end{subfigure}
\begin{subfigure}[]{0.48\columnwidth}
\centering
	\includegraphics[width=0.95\columnwidth]{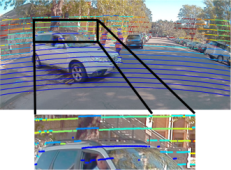}
    \caption{C masked camera image. }
    \label{sub2_a}
    \end{subfigure}
    
\begin{subfigure}[]{0.48\columnwidth}
\centering
	\includegraphics[width=0.95\columnwidth]{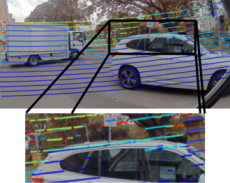}
    \caption{R camera image.}
    \label{sub2_b}
    \end{subfigure}
\begin{subfigure}[]{0.48\columnwidth}
\centering
	\includegraphics[width=0.95\columnwidth]{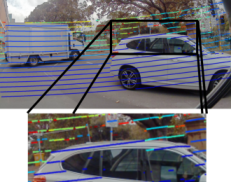}
    \caption{R masked camera image.}
    \label{sub2_c}
    \end{subfigure}
    
\begin{subfigure}[]{0.48\columnwidth}
\centering
	\includegraphics[width=0.95\columnwidth]{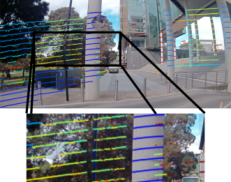}
    \caption{SR camera image. }
    \label{sub2_d}
    \end{subfigure}
\begin{subfigure}[]{0.48\columnwidth}
\centering
	\includegraphics[width=0.95\columnwidth]{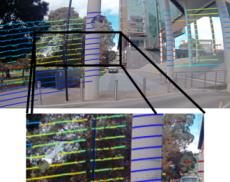}
    \caption{SR masked camera image.}
    \label{sub2_e}
    \end{subfigure}

\caption{\small Comparison between direct point cloud projection (first column) and with the extension of occlusion masking (second column) for each camera on the vehicle: left side (LS),  left (L), centre (C), right (R), right side (RS). The color of the projected point cloud varies with depth.}
\label{fig:masked_images}
\end{figure}

The mask shape corresponds to a rectangle where the dimensions $x_{Gap}$ and $y_{Gap}$ depend on the vertical and horizontal resolution of the lidar. The parameters selected for this paper are based on the VLP-16 documentation \cite{velodyne}. The gap between points (vertical or horizontal) can be calculated with the following equation:
\begin{equation} \label{eq12}
    Gap=d_t*\tan(\theta_{v-h})
\end{equation}
where $d_t$ is the distance to the target and $\theta_{v-h}$ is the vertical or horizontal angle between scan lines of consecutive points. In order to calculate the gaps in the pixels when the point cloud is projected, we make use of the generic camera-image projection equations \ref{eq_pc1} and \ref{eq_pc4}, obtaining:

\begin{align} \label{eq13}
u_{Gap} &= f_x \frac{d_t\tan(\theta_h)}{z} &   v_{Gap} &= f_y \frac{d_t\tan(\theta_v)}{z}
\end{align}

For our case we have assumed that the difference between $d_t$ and $z$ is negligible due to the adjacency of the lidar to the cameras. Therefore, $u_{Gap}$ and $v_{Gap}$ could be computed in terms of camera intrinsic parameters and lidar angular resolution:

\begin{align} \label{eq14}
u_{Gap} &= f_x \tan(\theta_h) & v_{Gap} &= f_y \tan(\theta_v)   
\end{align}

In the case of the lidar VLP-16 and our GMSL cameras, $\theta_v = 2 ^\circ$ and $\theta_h = 0.1 ^\circ$, giving $y_{Gap}= 41$ pixels and $x_{Gap} = 3$ pixels. Having calculated the size in pixels of the vertical and horizontal gaps, we proceed to project the ordered point cloud into the image, starting with the closest point to the camera. Each projected point will create a rectangular mask of $x_{Gap}$ and $y_{Gap}$ dimensions centred in the corresponding pixel. A more distant 3D point will not be included in the final pointcloud if its projection lies inside a masked area, in this case we assume the point is occluded. 
Fig. \ref{fig:masked_images} shows the difference between the lidar-image projection without and with occlusion masking for each camera.

\section{RESULTS}

We tested our algorithm using the Usyd Dataset \cite{usyd_dataset}. An electric vehicle equipped with multiple sensors was driven around the University of Sydney while collecting images (from five $100^\circ$ field of view cameras located around the vehicle), point cloud (from one lidar VLP-16 located on the roof) and odometry information. 

We assess the performance of the proposed pipeline using a single laser scan by implementing the lidar-image projection using three different methods: direct projection, projection after correcting the point cloud for motion and projection with occlusion handling after motion correction. 
A total of 20 single lidar scans were hand labelled for the experiment and then compared with the outcome of each method. Since the information projected into the point cloud corresponds to a semantic class probability distribution, the comparison is done between the ground truth and the semantic class with highest probability. 

We combined and discarded unused semantic classes obtaining 7 final labels in the point cloud (sky and unlabeled classes were discarded while pole and sign, pedestrian and rider and, building and fence were merged). Table \ref{table:SingleEvaluation} shows the results of the recall, precision and F1 score per experiment.

\begin{table*}[ht]
\centering
\vspace{3mm}
\caption{Single scan evaluation}
\label{table:SingleEvaluation}
\begin{tabular}{|c|c|c|c|c|c|c|c|c|c|}
\hline
\multirow{2}{*}{\textbf{Semantic class}} & \multicolumn{3}{c|}{\textbf{Direct Projection}} & \multicolumn{3}{c|}{\textbf{Projection + Motion Correction}} & \multicolumn{3}{c|}{\textbf{Projection + Motion C + Mask}} \\ \cline{2-10} 
                                & \textbf{Recall}  & \textbf{Precision } & \textbf{F1 Score} & \textbf{Recall}    & \textbf{Precision }    & \textbf{F1 Score}   & \textbf{Recall}  & \textbf{Precision }  & \textbf{F1 Score}  \\ \hline
Building                        & 0.749       & 0.785      & 0.769       & 0.769         & 0.804         & 0.786         & 0.809       & 0.849       & 0.829        \\ \hline
Pole                            & 0.702       & 0.172      & 0.276       & 0.722         & 0.185         & 0.295         & 0.731       & 0.215       & 0.332       \\ \hline
Road                            & 0.958       & 0.946      & 0.952       & 0.962         & 0.948         & 0.955         & 0.963       & 0.954       & 0.958        \\ \hline
Undrivable Road                         & 0.769       & 0.657      & 0.709       & 0.787         & 0.677         & 0.728         & 0.824       & 0.728       & 0.773        \\ \hline
Vegetation                      & 0.862       & 0.933      & 0.896       & 0.875         & 0.948         & 0.910         & 0.898       & 0.969       & 0.932        \\ \hline
Vehicle                         & 0.940       & 0.825      & 0.879       & 0.954         & 0.834         & 0.890         & 0.962       & 0.847       & 0.901        \\ \hline
Pedestrian                      & 0.938       & 0.361      & 0.521       & 0.953         & 0.375         & 0.538         & 0.987       & 0.649       & 0.783        \\ \hline
\end{tabular}
\end{table*}

The number of labelled points per scan is reduced by around 7$\%$ after incorporating the occlusion rejection process when compared to the original pointcloud. The recall is greatly improved when using the masking technique as it avoids the situation where occluded points were being mislabelled. The precision is also positively affected because of the reduction in the number of both false negatives and false positives.  
Similarly, the motion correction of the point cloud leads to the more accurate transfer of labels due to the correction of each point based on the vehicle position and image timestamp. 

From Table \ref{table:SingleEvaluation}, it is evident that the semantic classes with the most improvement in the metrics are pole and pedestrian. As poles are generally skinny and lidars generally have comparatively lower vertical resolution compared to horizontal resolution there are very few points covering these types of objects. If there is an error in the projection due to forwards or backwards motion of the vehicle, the transferred labels are likely to be incorrect. As pedestrians can be often considered as moving objects, the synchronisation between the point cloud and the image plays a fundamental role when combining the the sensors' information.

\begin{figure*}[h!]
\vspace{3mm}
\centering

\begin{subfigure}[]{\textwidth}
\centering
	\includegraphics[width=0.98\textwidth]{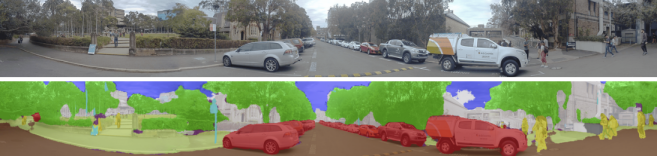}
    \caption{Stitched images for a 350$^\circ$ field of view of the surroundings. }
    \label{sub_es}
    \end{subfigure}

\begin{subfigure}[]{0.66\columnwidth}
\centering
	\includegraphics[width=0.9\columnwidth, height=3.5cm, trim={0 4cm 0  4cm},clip]{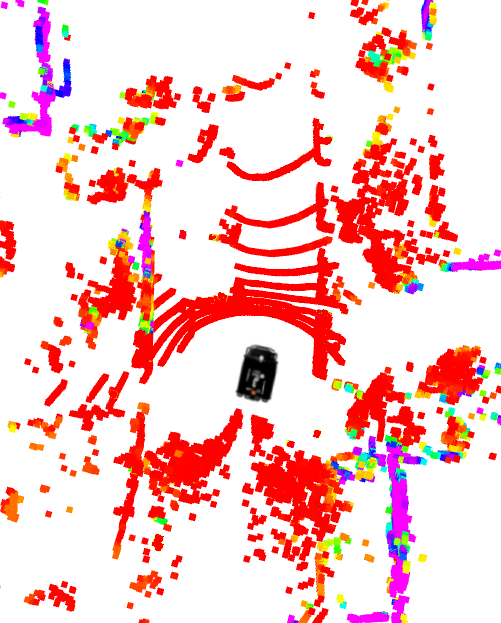}
    \caption{Buildings.}
    \label{sub_p_b}
    \end{subfigure}
\begin{subfigure}[]{0.66\columnwidth}
\centering
	\includegraphics[width=0.9\columnwidth, height=3.5cm, trim={0 4cm 0  4cm},clip]{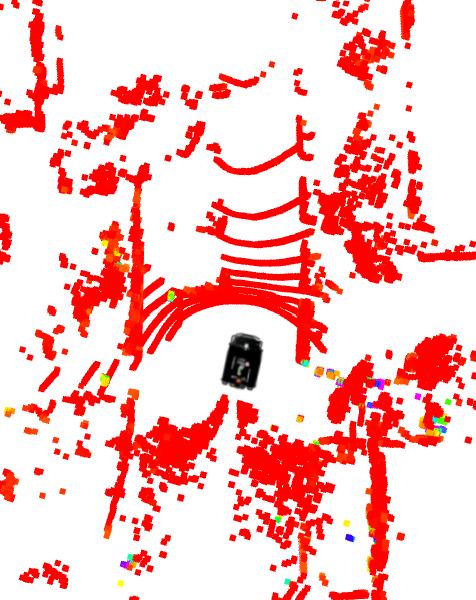}
    \caption{Poles and pedestrians. }
    \label{sub_p_pp}
    \end{subfigure}
\begin{subfigure}[]{0.66\columnwidth}
\centering
	\includegraphics[width=0.9\columnwidth, height=3.5cm, trim={0 4cm 0  4cm},clip]{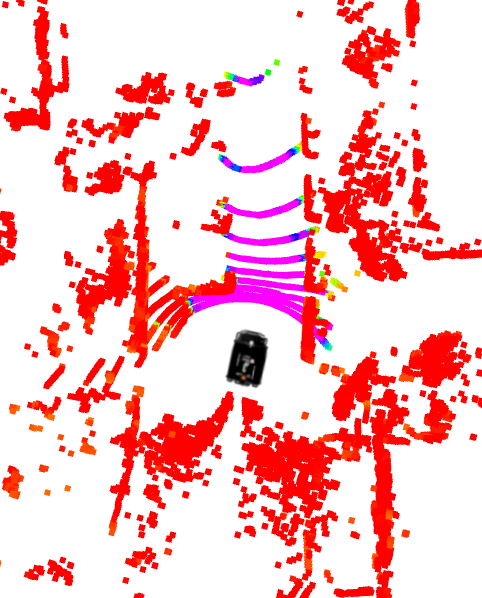}
    \caption{Road.}
    \label{sub_p_r}
    \end{subfigure}
    
\begin{subfigure}[]{0.66\columnwidth}
\centering
	\includegraphics[width=0.9\columnwidth, height=3.5cm, trim={0 4cm 0  4cm},clip]{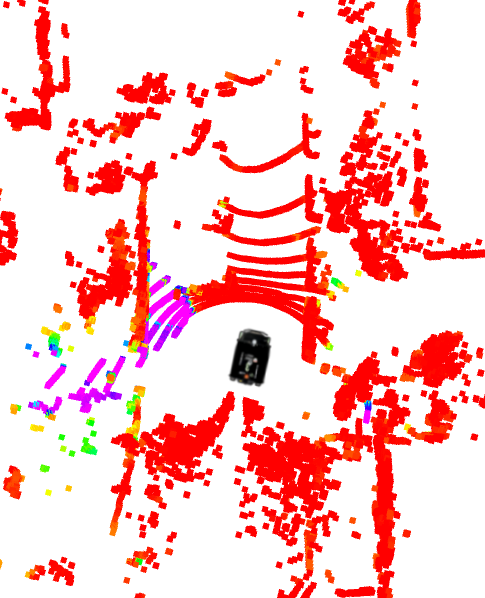}
    \caption{Undrivable road.}
    \label{sub_p_ur}
    \end{subfigure}
\begin{subfigure}[]{0.66\columnwidth}
\centering
	\includegraphics[width=0.9\columnwidth, height=3.5cm, trim={0 4cm 0  4cm},clip]{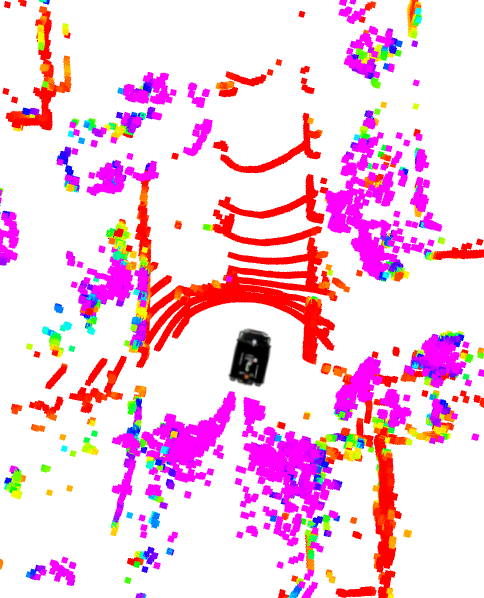}
    \caption{Vegetation. }
    \label{sub_p_v}
    \end{subfigure}
\begin{subfigure}[]{0.66\columnwidth}
\centering
	\includegraphics[width=0.9\columnwidth, height=3.5cm, trim={0 4cm 0  4cm},clip]{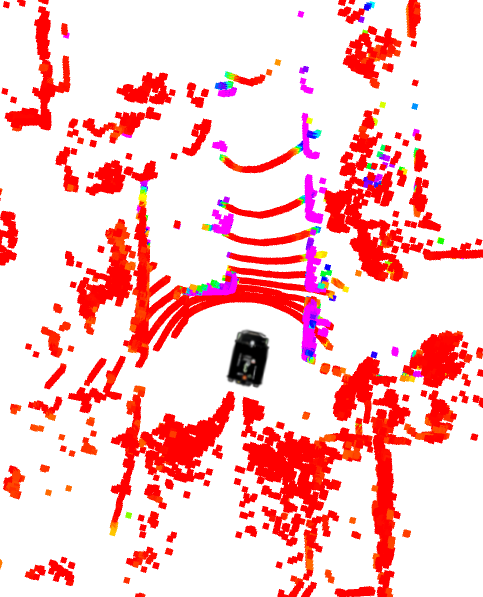}
    \caption{Vehicles.}
    \label{sub_p_vh}
    \end{subfigure}

\begin{subfigure}[]{\textwidth}
\centering
	\includegraphics[width=0.99\textwidth, trim={0 4cm 0  4cm},clip]{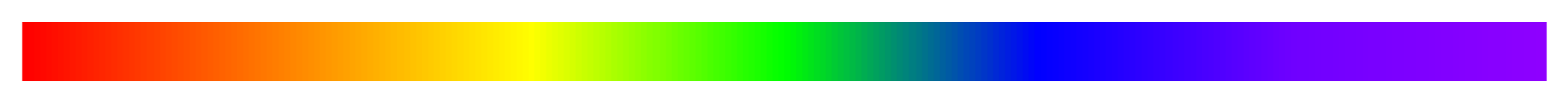}
    \end{subfigure}

\caption{\small \ref{sub_es} Semantic segmentation results from images taken by the vehicle of its surroundings. A single-scan point cloud from the 16 beam lidar is shown with probabilities displayed for each separate semantic class; the red color indicates the lowest probability $(0.0)$ and violet the highest $(1.0)$.  
In \ref{sub_p_b} is possible to see some walls in the scene indicated in purple; in the right-bottom of the image a 90$^\circ$ angle corresponding to a building corner is visible. In \ref{sub_p_pp}, scattered green, blue and purple points represent poles and pedestrians around the vehicle. In \ref{sub_p_r}, the road can be clearly seen directly in front of the vehicle. In \ref{sub_p_ur}, the portion of the footpath visible to both the lidar and camera can be seen classified as undrivable road. In the remaining images \ref{sub_p_v} and \ref{sub_p_vh}, vegetation and vehicles (on the side of the road) can be seen coloured in purple. 
}
\label{fig:six_uncert}
\end{figure*}

Due to the displacement between the lidar and camera, object classes closer to the platform occlude other structures which end up being incorrectly labelled as the class of the closest object, increasing the number of false positives. This in one of the reasons why the pole and pedestrian classes have the lowest precision when projected into the point cloud domain, and why this improves dramatically with the proposed occlusion handling method. 

Qualitative evaluation for a projection of semantic information into a single lidar scan is shown in Fig. \ref{fig:six_uncert}. In Fig. \ref{sub_es} we can observe the images of the five cameras stitched and semantically segmented. Fig. \ref{sub_p_b}-\ref{sub_p_v} depict a top-down view of the lidar data with the probability of each class shown in the point cloud domain. There is a strong correspondence between the high probability (in purple) and the object in the 3D world. 

One issue identified with this approach is that some objects only partially occlude the more distant objects. 
This can be seen in particular with the  vegetation class where even if the camera can see walls between tree branches, the CNN usually identifies this space as vegetation. As a result, points with a high probability of being vegetation can be found on building walls.


\section{CONCLUSIONS AND FUTURE WORK}

In this paper, we described an innovative approach to fuse information from different sensor modalities (cameras and lidar), resulting in a probabilistic semantic point cloud. 

Captured images were segmented by a previously trained CNN. Instead of using the output of the network directly, we opted to link heuristic probabilities to each class. These probabilities were calculated based on a superpixel segmentation of the raw image, the network output and score maps. The outcome of this process is the analytical probability for every semantic class per pixel.  
We presented an approach that takes a set of lidar packets with given timestamps and applies motion correction to the lidar packets given the vehicle odometry and a reference timestamp. Finally, the motion corrected lidar points are projected into the camera coordinate system. 

The results shown in Table \ref{table:SingleEvaluation} demonstrate the importance of correcting the lidar point cloud. The precision of the projection improves significantly due to this process as it improves the corespondance between the points and the image. 
This improvement implies a more precise transference of information and hence more true positives and less false positives.
 
To project accurate information from the image to the lidar, we presented an algorithm to determine if each lidar 3D point can be seen by each camera. This algorithm is essential to remove occluded points that are visible to the lidar but that the camera cannot see.  
We have proposed a masking methodology to cope with the occlusion problem for the VLP-16 lidar, which was unable to be solved using traditional techniques due to the low vertical resolution. 
This approach rejects a number of points that are primarily false positives or false negatives. 
This methodology can be easily extended to being used with different lidar-camera arrangements. 

The results of this paper are relevant in terms of semantic sensor fusion (with labels heuristic probabilities) while coping with issues as motion correction and occlusions. As well, the resulting point cloud can be used in a number of applications where current probabilistic semantic information is needed to make decisions or to build semantic maps of the environment. 

As future work, we intend to propagate the odometry uncertainty into the projection process and integrate it into the pipeline, and evaluate the registration of the resulting point cloud. 

\section*{ACKNOWLEDGMENT}

This work has been funded by the ACFR, the University of Sydney through the Dean of Engineering and Information Technologies PhD Scholarship (South America) and the Australian Research Council Discovery Grant DP160104081 and University of Michigan / Ford Motors Company Contract ``Next generation Vehicles".

\bibliography{main}
\bibliographystyle{IEEEtran}

\end{document}